 \documentclass[pmlr,twocolumn,10pt]{jmlr} % W&CP article

\usepackage{booktabs}
\usepackage{siunitx}
\usepackage{float}
\usepackage{multirow}
\usepackage[switch]{lineno}

\theorembodyfont{\upshape}
\theoremheaderfont{\scshape}
\theorempostheader{:}
\theoremsep{\newline}

\jmlrvolume{259}
\jmlryear{2024}
\jmlrpublished{LEAVE UNSET}
\jmlrworkshop{Machine Learning for Health (ML4H) 2024} % W&CP title

\title{Exploring Complex Mental Health Symptoms via Classifying Social Media Data with Explainable LLMs}

  \author{%
   \Name{Kexin Chen} \Email{co.chen@mail.utoronto.ca}\\
   \addr University of Toronto, Canada \\
   \Name{Noelle Lim} \Email{arinlim@gmail.com} \\
   \addr LinkedIn Corporation, USA \\
   \Name{Claire Lee} \Email{skclairelee@gmail.com} \\
   \addr Princeton University, USA \\
   \Name{Michael Guerzhoy} \Email{guerzhoy@cs.toronto.edu}\\
   \addr University of Toronto, Canada\\ 
}

\begin{document}

\maketitle

\begin{abstract}
We propose a pipeline for gaining insights into complex diseases by training LLMs on challenging social media text data classification tasks, obtaining explanations for the classification outputs, and performing qualitative and quantitative analysis on the explanations. We report initial results on predicting, explaining, and systematizing the explanations of predicted reports on mental health concerns in people reporting Lyme disease concerns. We report initial results on predicting future ADHD concerns for people reporting anxiety disorder concerns, and demonstrate preliminary results on visualizing the explanations for predicting that a person with anxiety concerns will in the future have ADHD concerns.
\end{abstract}

% \paragraph*{Data and Code Availability}
% The paper uses data from Reddit, including:  all posts containing the keyword \texttt{"lyme"},  posts from \texttt{/r/Anxiety} and \texttt{/r/ADHD} subreddits, and posts from the \texttt{/r/Askdocs} subreddit. Scripts for obtaining the data will be available but are not currently available. Code will be available but is not available currently.

% \paragraph*{Institutional Review Board (IRB)}
% IRB information will be provided.

\section{Introduction}
% \label{sec:intro}

A substantial proportion of people with mental health issues do not consult a specialist~\citep{statcan2023mentalhealth} and many turn online to discuss their concerns~\citep{giles2011self}. For many complex diseases, but especially for diseases with mental health aspects, syndromes interact with the nosology\footnote{nosology(n): the branch of medical science dealing with the classification of diseases.
} and the cultures of the clinician specializations in complex and ill-understood ways~\citep{zachar2017philosophy}. Social media offers a wealth of data not traditionally available to clinicians, with millions of patients talking about their symptoms~\citep{low2020natural}.

In this paper, we propose a pipeline for gaining new insights from large-scale social media data:

\begin{enumerate}
    \item Identify a challenging classification task
    \item Learn LLM-based classifiers for the task
    \item Generate explanations for the LLM's classification outputs
    \item Use quantitative and qualitative analyses to gain new insights from the data
\end{enumerate}

In this work, we apply this approach to two tasks. 
First, building on the work in~\citep{lee2024detecting}, we classify social media users in the \texttt{r/anxiety} subreddit who have not yet posted in the \texttt{r/ADHD} subreddit as ones that will or will not post in the \texttt{r/ADHD} subreddit in the future, and then visualize the explanations for that classifier. The task is interesting because it can be used as a proxy task for detecting ADHD concerns in people with anxiety concerns, because the task seems impossible for keyword-based architectures~\citep{lee2024detecting}, and the visualization of prediction explanations in this task can lead in to insights about both symptoms and nosology of anxiety disorders and ADHD.
Second, we classify Reddit posts that potentially contain concerns about Lyme disease as possibly containing mental health concerns related to Lyme or not containing mental health related to Lyme, based on a small manually-labelled dataset. The task is interesting because it allows us to iteratively obtain a larger dataset of social media posts related to both Lyme disease and mental health. We analyze the explanations for the classifier that classifies posts as mental health-related or not in order to obtain possibly-novel preliminary insights into the mental health aspects of Lyme disease.

\section{Background}
%\label{sec:intro}

Research shows that up to 53\% of adults with ADHD may also suffer from an anxiety disorder, while approximately 28\% of adults with an anxiety disorder may have undiagnosed ADHD~\citep{quinn2014review}~\citep{van2011adult}. This overlap can complicate clinical decision-making, as anxiety or depression symptoms are often treated without recognizing the potential presence of ADHD, resulting in only the symptoms for anxiety being treated and ADHD beign left untreated~\citep{katzman2017adult}. 
Lyme disease is known for its significant neurological and psychological impacts. Approximately 40\% of individuals diagnosed with Lyme disease experience neurological symptoms, with a notable percentage also developing mental health disorders~\citep{fallon1994lyme}. The prevalence of mental health disorders for those diagnosed with Lyme disease is much higher than in the general population~\citep{fallon2021lyme}. 

\section{Methods}
%\label{sec:intro}

\subsection{Task Description}
\subsubsection{Will exclusive \texttt{/r/Anxiety} poster also post in \texttt{/r/ADHD} in the future?}

In~\cite{lee2024detecting}, data is collected from Reddit to classify posts from users who only discuss anxiety in the \texttt{/r/Anxiety} subreddit versus those who transition from posting in the \texttt{/r/Anxiety} to discussing ADHD in \texttt{/r/Anxiety}. This classification task serves as a proxy for identifying individuals who might be developing awareness of their potential comorbid ADHD condition. 

\subsubsection{Mental health aspects of Lyme disease}

We classify Lyme disease-related posts as either mental health-related or not using RoBERTa models, which have demonstrated their utility in
classifying mental health disorders from textual data on social media~\citep{malviya2021transformers}~\citep{murarka2021classification}~\citep{lee2024detecting}.
\subsection{Datasets}
\paragraph{ADHD and Anxiety Dataset} Text data is collected from the \texttt{/r/Anxiety} and \texttt{/r/ADHD}
subreddits~\citep{lee2024detecting}. Posts are included if users exclusively post in the \texttt{/r/Anxiety} or initially posted there before starting to also post in   \texttt{/r/ADHD}, with a total of 47,482 posts collected from before 2023. Posts from \texttt{/r/ADHD} are excluded if posted within six months of the first \texttt{/r/Anxiety} post. This results in a total of 47,482 posts, of which 33\% are retained for testing purposes.

\paragraph{Manually-labelled Mental-Health-Related\\ Symptoms of Lyme disease Dataset}

We collect a dataset of 58,398 Reddit posts containing the term ``Lyme" from Reddit from before 2022. A subset of 343 posts is manually labeled as mental-health-related or non-health-related based on specific criteria. Any of the following are reasons to label a post as mental-health-related: 1) Mention of specific mental health symptoms 2) Ideation of self-harm or suicide 3) emotional distress 4) anger management issues 5) social relationship challenges 6) persistent feelings of emptiness 7) Substance use issues. 

% \begin{table*}[!h]
% \floatconts
%   {tab:combined_results}%
%   {\caption{Performance metrics for predicting mental-health-related posts in ADHD and Lyme datasets (balanced for 50\% baserate)}}%
%   {%
% \begin{tabular}{llccc}
% \toprule
% \textbf{Dataset} & \textbf{Metric} & \textbf{Logistic Regression} & \textbf{Naïve Bayes} & \textbf{RoBERTa} \\
% \midrule
% \multirow{2}{*}{Will post in ADHD after posting in Anxiety} 
% & Accuracy   & 0.48 & 0.59 & 0.76 \\
% & F1 Score   &  0.61   &   0.59   &   /   \\
% \midrule
% \multirow{2}{*}{Lyme} 
% & Accuracy   & 0.81 & 0.72 & 0.88 \\
% & F1 Score   & 0.42 & 0.47 & 0.73 \\
% \bottomrule
% \end{tabular}
% }
% \end{table*}
\begin{table*}[!h]
\floatconts
  {tab:combined_results}%
  {\caption{Performance metrics for predicting mental-health-related posts in ADHD and Lyme datasets (balanced for 50\% baserate)}}%
  {%
\resizebox{\textwidth}{!}{%
\begin{tabular}{llccc}
\toprule
\textbf{Dataset} & \textbf{Metric} & \textbf{Logistic Regression} & \textbf{Naïve Bayes} & \textbf{RoBERTa} \\
\midrule
\multirow{2}{*}{Will post in ADHD after posting in Anxiety} 
& Accuracy   & 0.48 & 0.59 & 0.89 \\
& F1 Score   &  0.61   &   0.59   &   0.89   \\
\midrule
\multirow{2}{*}{Lyme} 
& Accuracy   & 0.81 & 0.72 & 0.88 \\
& F1 Score   & 0.42 & 0.47 & 0.73 \\
\bottomrule
\end{tabular}%
}
}
\end{table*}

\begin{table*}[!h]
\floatconts
  {tab:mask-examples}%
  {\caption{Examples of RoBERTa classifier's explanation for positive predictions for ADHD (ADHD/Anxiety) and Mental Health concerns (Lyme) on Reddit posts rephrased and altered for anonymity}}%
  {%
\begin{tabular}{p{0.075\textwidth} p{0.875\textwidth}}
    \toprule
    \textbf{Datasets} & \textbf{Examples} \\
    \midrule
    
    \textbf{ADHD} \newline \textbf{and} \newline \textbf{Anxiety} & \textbf{Post:} \textcolor{red}{I missed a deadline by several days and I’m extremely stressed. I haven't felt well over the past weekend.} I've had multiple arguments with my mom, and our relationship is more strained than ever. \textcolor{red}{I ended up procrastinating and avoided urgent tasks, including reading the email about the assignment I missed.} I finally checked it today and realized I've been behind for days. I know I should explain the situation to my instructor and complete the work as soon as possible, but the consequences feel significant this time, and I can't stop worrying about it. I also don't know how to explain my situation without it sounding like I'm making excuses. I’m lost. I can't even focus on the email to understand what I need to do now; my eyes literally hurt from reading it. \\
    \midrule
    
    \textbf{Lyme} \newline \textbf{and} \newline \textbf{Mental Health} & \textbf{Post:} Five years ago, I contracted Lyme disease and experienced something quite unusual and intriguing; \textcolor{red}{my brain felt like it was constantly operating at full capacity.} It's hard to put the sensation into words, \textcolor{red}{but I recall it involved a struggle to gauge size (everything appeared enormous yet proportional to everything else, including myself),} which made me constantly attempt to "render" objects in my mind without any success. \textcolor{red}{This phenomenon only occurred at night and was particularly noticeable when I tried to fall asleep and began to lightly dream before dozing off.} On the nights this occurred, I would be in a delirious state in all but one instance (total of 5-6 episodes). Although these episodes began when I got Lyme disease, they continued sporadically afterward, with the last one occurring about two years after the initial infection, and were usually, though not always, linked to a fever. Is there a name for this phenomenon? Where can I find more information about it? \\
    
    \bottomrule
  \end{tabular}
}
\end{table*}

\begin{table*}[!h]
\floatconts
  {tab:phrase-frequency-table}%
  {\caption{Comparison of counts of unusual symptoms that we also highlighted by RoBERTa as explanation for classifying post as mental-health-related in  the Lyme dataset (out of top 300 matches) and AskDocs (reference) (out of top 300) dataset}}%
  {%
% \begin{tabular}{p{0.6\textwidth} p{0.15\textwidth} p{0.15\textwidth}}
%     \toprule
%     \textbf{Phrases} & \textbf{Lyme Dataset} & \textbf{AskDocs (reference)} \\
%     \midrule
%     The ceiling of my current apartment has mold, and the landlord is slow to clean it up [theme: mold] & 132 & 59 \\
%     \midrule
%     I experience sinus pressure that feels like it's squeezing my nose, with tension noticeable when I look downward [theme: sinus/nose] & 68 & 54 \\
%     \midrule
%     My weight issues stem from a combination of an eating disorder and thyroid problems [theme: weight/eating disorder/thyroid] & 15 & 10 \\
%     \midrule
%     My body clenches and contracts involuntarily, with intense abdominal twitches causing my body to jerk [theme: clenching/contractions] & 136 & 125 \\
%     \midrule
%     Even now, I have flare-ups when highly stressed, mainly manifesting as persistent facial pressure [theme: flare-ups] & 30 & 24 \\
%     \midrule
%     %She experiences episodes of unsteady walking and involuntary head movements & 5 & 9 \\
%     %\midrule
%     I suddenly developed significant cross-eyed vision [theme: vision] & 16 & 12 \\
%     \midrule
%     % I notice strange visual artifacts, like jumping effects, when looking at high-contrast patterns (such as checkerboards) & 2 & 2 \\
%     % \midrule
%     % I get "brain zaps" where I'm jolted awake by loud, cymbal-like sounds in my head & 2 & 2 \\
%     \bottomrule
%   \end{tabular}
\begin{tabular}{p{0.6\textwidth} p{0.15\textwidth} p{0.15\textwidth}}
    \toprule
    \textbf{Phrases} & \textbf{Lyme Dataset (\%)} & \textbf{AskDocs (reference)} \\
    \midrule
    The ceiling of my current apartment has mold, and the landlord is slow to clean it up [theme: mold] & 44.0 & 19.7 \\
    \midrule
    I experience sinus pressure that feels like it's squeezing my nose, with tension noticeable when I look downward [theme: sinus/nose] & 22.7 & 18.0 \\
    \midrule
    My weight issues stem from a combination of an eating disorder and thyroid problems [theme: weight/eating disorder/thyroid] & 5.0 & 3.3 \\
    \midrule
    My body clenches and contracts involuntarily, with intense abdominal twitches causing my body to jerk [theme: clenching/contractions] & 45.3 & 41.7 \\
    \midrule
    Even now, I have flare-ups when highly stressed, mainly manifesting as persistent facial pressure [theme: flare-ups] & 10.0 & 8.0 \\
    \midrule
    %She experiences episodes of unsteady walking and involuntary head movements & 5 & 9 \\
    %\midrule
    I suddenly developed significant cross-eyed vision [theme: vision] & 5.33 & 4.0 \\
    \midrule
    % I notice strange visual artifacts, like jumping effects, when looking at high-contrast patterns (such as checkerboards) & 2 & 2 \\
    % \midrule
    % I get "brain zaps" where I'm jolted awake by loud, cymbal-like sounds in my head & 2 & 2 \\
    \bottomrule
  \end{tabular}
}
\end{table*}

\subsection{Expanded Mental-Health-Related Symptoms of Lyme Dataset}

We train a RoBERTA classifier to predict whether a post is mental-health-related or not, and classify 58,398 posts containing the keyword ``Lyme" as mental-health-related or not. We retain the potentially mental-health-related Lyme posts.

\subsection{Baseline AskDocs dataset}
We scrape the latest 5,000 posts from before 2022 from the general-medical-interest \texttt{r/AskDocs} subreddit as a reference set.

\subsection{Classification results}

For the mental-health-related datasets, we present results with baselines as well as with a fine-tuned RoBERTa  using a 80/20 training/test split. The results are shown in~\tableref{tab:combined_results}. 

\subsection{Generating explanations}
%\label{sec:intro}

We generate explanations as follows for the RoBERTa classifier: each post is broken up into phrases (defined as sentence fragments enclosed in punctuation marks). Each phrase in turn is masked out, and the phrases that influence the classifier output the most are highlighted. The examples of RoBERTa classifier’s highlights on Reddit posts for mental health indicators are shown in \tableref{tab:mask-examples},  and more examples for Lyme disease posts related to mental health are presented in \tableref{tab:appendix-single-column-examples} in the Appendix. All the examples we show in the paper have been rephrased to ensure that user privacy is protected.

\subsection{Using Explanations to Explore Mental Health Aspects of Lyme Disease}
%\label{sec:intro}

We use the following procedure to explore the mental health aspects of Lyme disease.

\begin{enumerate}
    \item Classify 5,000 posts containing ``Lyme" as mental-health related or not
    \item Obtain explanations for the posts classified as mental-health-related
    \item Iterate by grouping the explanations together manually
    \item Obtain OpenAI \texttt{text-embedding-3-large} embeddings for each group of explanations and manually look through the top 300 matches by cosine similarity in all the Lyme posts for more similar explanations
    \item Look through the top 300 matches for each group of explanations in the reference \texttt{r/AskDocs} corpus.
\end{enumerate}

Preliminary results are available in \tableref{tab:phrase-frequency-table}. We identify that the mold theme appears more prevalent in the Lyme dataset than in the baseline AskDocs dataset (p-value $< 0.01$ when using a simple random sample model) in the hand-labelled top 300 matches for the symptom description, indicating a potential avenue to explore in the three-way interaction of expressing concern about Lyme, expressing mental-health concerns, and speaking about mold.

\section{Conclusion}
We work with two mental-health-related social media text datasets. We demonstrate a pipeline of identifying an interesting and challenging classification task, obtaining explanations for RoBERTa's classification decisions on the tasks, and exploring those explanations. For the Lyme dataset, we demonstrate preliminary results showing a proof-of-concept of a pipeline where the explanations of the classifier are used in order to identify interesting patterns in the dataset.

\bibliography{references}

\begin{thebibliography}{12}
\providecommand{\natexlab}[1]{#1}
\providecommand{\url}[1]{\texttt{#1}}
\expandafter\ifx\csname urlstyle\endcsname\relax
  \providecommand{\doi}[1]{doi: #1}\else
  \providecommand{\doi}{doi: \begingroup \urlstyle{rm}\Url}\fi

\bibitem[Ameringen et~al.(2011)Ameringen, Mancini, Simpson, and Patterson]{van2011adult}
Michael~Van Ameringen, Catherine Mancini, William Simpson, and Beth Patterson.
\newblock Adult attention deficit hyperactivity disorder in an anxiety disorders population.
\newblock \emph{CNS Neurosci. Ther.}, 17\penalty0 (4):\penalty0 221--226, April 2011.

\bibitem[Fallon and Nields(1994)]{fallon1994lyme}
Brian~A. Fallon and Jenifer~A. Nields.
\newblock Lyme disease: A neuropsychiatric illness.
\newblock \emph{Am. J. Psychiatry}, 151\penalty0 (11):\penalty0 1571--1583, November 1994.

\bibitem[Fallon et~al.(2021)Fallon, Madsen, Erlangsen, and Benros]{fallon2021lyme}
Brian~A. Fallon, Trine Madsen, Annette Erlangsen, and Michael~E. Benros.
\newblock Lyme borreliosis and associations with mental disorders and suicidal behavior: A nationwide danish cohort study.
\newblock \emph{Am. J. Psychiatry}, 178\penalty0 (10):\penalty0 921--931, October 2021.

\bibitem[Giles and Newbold(2011)]{giles2011self}
David~C. Giles and Julie Newbold.
\newblock Self-and other-diagnosis in user-led mental health online communities.
\newblock \emph{Qual. Health Res.}, 21\penalty0 (3):\penalty0 419--428, March 2011.

\bibitem[Katzman et~al.(2017)Katzman, Bilkey, Chokka, Fallu, and Klassen]{katzman2017adult}
Martin~A. Katzman, Terrence~S. Bilkey, Pratap~R. Chokka, Angelo Fallu, and Laureen~J. Klassen.
\newblock Adult adhd and comorbid disorders: Clinical implications of a dimensional approach.
\newblock \emph{BMC Psychiatry}, 17\penalty0 (1), July 2017.
\newblock \doi{10.1186/s12888-017-1463-3}.

\bibitem[Lee et~al.(2024)Lee, Lim, and Guerzhoy]{lee2024detecting}
Claire~S. Lee, Noelle Lim, and Michael Guerzhoy.
\newblock Detecting a proxy for potential comorbid adhd in people reporting anxiety symptoms from social media data.
\newblock In \emph{9th Workshop on Computational Linguistics and Clinical Psychology}, page 172, 2024.

\bibitem[Low et~al.(2020)Low, Rumker, Talkar, Torous, Cecchi, and Ghosh]{low2020natural}
Daniel~M. Low, Laurie Rumker, Tanya Talkar, John Torous, Guillermo Cecchi, and Soumya~S. Ghosh.
\newblock Natural language processing reveals vulnerable mental health support groups and heightened health anxiety on reddit during covid-19: Observational study.
\newblock \emph{J. Med. Internet Res.}, 22\penalty0 (10):\penalty0 e22635, October 2020.
\newblock \doi{10.2196/22635}.

\bibitem[Malviya et~al.(2021)Malviya, Roy, and Saritha]{malviya2021transformers}
Keshu Malviya, Bholanath Roy, and S.K. Saritha.
\newblock A transformers approach to detect depression in social media.
\newblock In \emph{2021 International Conference on Artificial Intelligence and Smart Systems (ICAIS)}, pages 718--723. IEEE, March 2021.

\bibitem[Murarka et~al.(2021)Murarka, Radhakrishnan, and Ravichandran]{murarka2021classification}
Ankit Murarka, Balaji Radhakrishnan, and Sushma Ravichandran.
\newblock Classification of mental illnesses on social media using roberta.
\newblock In \emph{Proceedings of the 12th International Workshop on Health Text Mining and Information Analysis}, pages 59--68, 2021.

\bibitem[Quinn and Madhoo(2014)]{quinn2014review}
Patricia~O. Quinn and Manisha Madhoo.
\newblock A review of attention-deficit/hyperactivity disorder in women and girls.
\newblock \emph{Prim. Care Companion CNS Disord.}, 16\penalty0 (3), June 2014.
\newblock \doi{10.4088/pcc.13r01596}.

\bibitem[{Statistics Canada}(2023)]{statcan2023mentalhealth}
{Statistics Canada}.
\newblock Mental health service utilization in canada, 2023.
\newblock URL \url{https://www150.statcan.gc.ca/n1/pub/75-006-x/2023001/article/00011-eng.htm}.
\newblock Accessed: 2024-08-29.

\bibitem[Zachar and Kendler(2017)]{zachar2017philosophy}
Peter Zachar and Kenneth~S. Kendler.
\newblock The philosophy of nosology.
\newblock \emph{Annu. Rev. Clin. Psychol.}, 13\penalty0 (1):\penalty0 49--71, May 2017.

\end{thebibliography}

\appendix

\section{Explanations from r/Lyme\label{apd:first}}

\begin{table*}[!h]
\floatconts
  {tab:appendix-single-column-examples}%
  {\caption{Examples of RoBERTa classifier's explanation for positive predictions for Mental Health concerns in Lyme-related Reddit posts, rephrased for privacy}}%
  {%
\begin{tabular}{p{0.95\textwidth}}
    \toprule
    \textbf{Examples} \\
    \midrule
    
    \textbf{Post:} my partner has been dealing with lyme since around 2015. \textcolor{red}{he has a hard time managing his hip pain,} neck and back pain, \textcolor{red}{persistent headaches that disturb his sleep, difficulty sleeping due to anxiety, and overall heightened anxiety levels.} i've never met someone with lyme before. i want to support him but need advice on how friends or family can best help someone with lyme. update: just got dumped for his ex, lol. \\
    \midrule
    
    \textbf{Post:} i've recently experienced a number of strange, debilitating symptoms. \textcolor{red}{it began with a stiff neck in late november, and in the past week, it has rapidly worsened to include headaches, severe depression,} a sense of unreality, hand tremors, \textcolor{red}{poor coordination,} no appetite or thirst, visual disturbances, and feeling extremely hot or cold in different parts of my body. \\
    \midrule
    
    \textbf{Post:} \textcolor{red}{i had personality disorders before contracting lyme, and it's interfering with my recovery,} i feel completely empty and unmotivated, \textcolor{red}{struggling with dependency issues, wanting to be taken care of, and lacking any willpower.} healing from lyme would be easier if my mental health wasn't such an obstacle. sometimes i end up not eating or falling into depression, losing the will to get up in the morning. i can't even commit to taking supplements or changing my diet anymore. i'm afraid my mental health issues might be my downfall. \\
    \midrule

    \textbf{Post:} \textcolor{red}{derealization and depersonalization (dpdr) feels like being intoxicated or out of reality,} floating sensations, \textcolor{red}{a feeling of disconnection from your body, thoughts, or words.} you don't recognize yourself or loved ones anymore. it's a terrifying experience. it feels like your brain is permanently altered, and \textcolor{red}{it seems like it will never return to normal.} has anyone here successfully treated this symptom? i have lyme and other co-infections, confirmed by cdc tests, and am five months into antibiotic treatment. \\
    \midrule

    \textbf{Post:} january 23: \textcolor{red}{sensation of bugs crawling on my scalp, had a paranoid episode thinking i had lice when i didn't. persistent crawling sensation led to paranoia, depression, and brain fog, forgetfulness with words. later that january: discovered a bump on my scalp, scratched it off, and it bled excessively. the area around it had a bullseye appearance. never saw a tick but had continuous crawling sensations. february 23: teeth grinding, jaw locking. thought it was related to my adhd meds, brain fog,} forgetfulness, attributed to adhd. menstrual cycle became longer, around 30-32 days. march 23: hospital admission for unexplained chest pain. abnormal blood tests suggested a possible infection but was sent home. cycle was again irregular, 32-33 days. april 23: noticed significant cognitive decline, increased adhd medication dosage. difficulty finding words, staying up later, restlessness, poor sleep quality. may 23: severe foot pain in the morning, difficulty walking, tight calf muscles, intense pms symptoms, extended cycle to 36 days, elevated liver enzymes. \\
    \midrule

    \textbf{Post:} hi everyone. i've been battling Lyme disease for over four years. \textcolor{red}{my pain has been particularly severe lately, disrupting my sleep and worsening my mood. it's making me less active and exacerbating my erratic sleep schedule.} what pain management strategies or sleep aids have worked well for you? pain medications aren't effective, and heat pads aren't helping. i'm at a loss, any suggestions? \\
    \midrule

    \textbf{Post:} \textcolor{red}{have you experienced anhedonia due to lyme?} to clarify, i don't mean general depression or reduced enjoyment, but a complete inability to feel pleasure or love for others. total numbness. \textcolor{red}{i rarely see discussions about this symptom, and when i do, it often sounds like it's mistaken for major depression.} total flatness and emotional numbness. has anyone else experienced this? how long did it last? \\
    
    \bottomrule
  \end{tabular}
}
\end{table*}

\end{document}